\documentclass[journal, letterpaper]{IEEEtran}

\usepackage{url}        

\usepackage{textgreek}	% Greek to me, dawg
\usepackage{listings}
\usepackage{csvsimple}
\usepackage{longtable}
\usepackage{hyperref}
\usepackage{booktabs}
\usepackage{amsfonts}
\usepackage{nicefrac}
\usepackage{microtype}
\usepackage{graphicx}
\usepackage{multicol}
\usepackage{multirow}
\usepackage{wrapfig}
\usepackage{bm}
\usepackage{tabularx}
\usepackage{cite}
\usepackage{amsmath,amssymb,amsfonts}
\usepackage{xcolor}
\usepackage{etoolbox}
\usepackage{mathtools}
\usepackage{hhline}
\usepackage{adjustbox}
\usepackage{array}
\usepackage{caption}
\usepackage{fancyhdr}

\begin{document}
\bstctlcite{IEEEexample:BSTcontrol}

\title{Identifying Multi-modal Knowledge Neurons in Pretrained Transformers via Two-stage Filtering}

\author{
    \begin{minipage}{0.1\textwidth}
        \centering
        \hspace{5mm}
    \end{minipage}
    \begin{minipage}{0.39\textwidth}
        \centering
        \textbf{Yugen Sato}\\
        Meiji University\\
        Kanagawa, Japan\\
        \texttt{yugen0104.tu@gmail.com}
    \end{minipage}
    \hfill
    \begin{minipage}{0.39\textwidth}
        \centering
        \textbf{Tomohiro Takagi}\\
        Meiji University\\
        Kanagawa, Japan\\
        \texttt{takagit@gmail.com}
    \end{minipage}
    \begin{minipage}{0.1\textwidth}
        \centering
        \hspace{5mm}
    \end{minipage}
}

\maketitle

\begin{abstract}
Recent advances in large language models (LLMs) have led to the development of multimodal LLMs (MLLMs) in the fields of natural language processing (NLP) and computer vision. Although these models allow for integrated visual and language understanding, they present challenges such as opaque internal processing and the generation of hallucinations and misinformation. Therefore, there is a need for a method to clarify the location of knowledge in MLLMs.

In this study, we propose a method to identify neurons associated with specific knowledge using MiniGPT-4, a Transformer-based MLLM. Specifically, we extract knowledge neurons through two stages: activation differences filtering using inpainting and gradient-based filtering using GradCAM. Experiments on the image caption generation task using the MS COCO 2017 dataset, BLEU, ROUGE, and BERTScore quantitative evaluation, and qualitative evaluation using an activation heatmap showed that our method is able to locate knowledge with higher accuracy than existing methods.

This study contributes to the visualization and explainability of knowledge in MLLMs and shows the potential for future knowledge editing and control.
\end{abstract}

% TODO: 図や表のタイトルなどを英訳
\section{Introduction}
In recent years, the advancement of large language models (LLMs) has established them as foundational models for numerous applications in the field of natural language processing (NLP). Inspired by this success, researchers in computer vision have extended the input modality of language models, leading to the development of multimodal large language models (MLLMs) capable of processing both text and images. MLLMs provide an integrated understanding of vision and language, achieving remarkable performance improvements in tasks such as image captioning and visual question answering (VQA).

One of the key underlying technologies contributing to the success of LLMs and MLLMs is the Transformer \cite{vaswani2017attention}. While the Transformer has revolutionized machine learning and contributed to performance improvements across various fields, its internal processes remain largely opaque, limiting its interpretability. In particular, for text generation models such as LLMs and MLLMs, this opacity can lead to unintended outputs, including the generation of harmful or misleading content \cite{gehman-etal-2020-realtoxicityprompts} and hallucinations \cite{weidinger2021ethical}. These issues highlight the urgent need to enhance model interpretability, not only for a deeper understanding of these models but also for ensuring responsible and ethical applications.

Improving interpretability would enable the identification and localization of knowledge within the Transformer, allowing for the modification or removal of specific knowledge to facilitate controlled model development and adaptation. This flexibility would significantly contribute to the refinement and responsible use of Transformer-based models.

In this study, as a first step toward enhancing the interpretability of Transformer-based text generation models, we conduct neuron-level observations using MiniGPT-4 \cite{zhu2023minigpt4enhancingvisionlanguageunderstanding} to clarify the localization of knowledge.

Several studies have approached similar challenges \cite{schwettmann2023multimodalneuronspretrainedtextonly, pan2024findingeditingmultimodalneurons, huang2024minerminingunderlyingpattern}. Research on MLLMs often emphasizes neurons that bridge textual and visual features. Notably, Schwettmann et al. \cite{schwettmann2023multimodalneuronspretrainedtextonly} extended this concept by identifying multimodal neurons even in purely textual components of LLMs, revealing their ability to bridge visual and textual representations. Additionally, Pan et al. \cite{pan2024findingeditingmultimodalneurons} proposed a method that scores neurons based on sensitivity, specificity, and causal effects, focusing on the final token of textual input to identify multimodal neurons associated with specific visual knowledge.

However, their approach primarily analyzes the influence of language tokens and does not directly elucidate how MLLMs internally retain and process knowledge through visual features. Existing methods rely on model-driven techniques based on internal representations and weights, while data-driven approaches that observe internal representations under varying input conditions remain largely unexplored.

To address this gap, we propose a data-driven approach that leverages the multimodal nature of MLLMs by analyzing neuron activations and gradients at image token positions within the input sequence. To demonstrate that the identified neurons are associated with target knowledge, we conduct text generation experiments with noise perturbation using the MS COCO dataset \cite{lin2015microsoftcococommonobjects}.

Our contributions can be summarized as follows:
\begin{itemize}
    \item We propose a novel method for identifying knowledge neurons in Transformer-based MLLMs.
    \item Through image caption generation experiments with noise perturbation and additional analyses, we demonstrate that the identified neurons are related to specific target knowledge.
\end{itemize}

\section{Method}
In this study, we first describe the architecture and model of the MLLMs used (§\ref{subsec:mllm}), then define neurons in the Transformer (§\ref{sec:Neurons in Transformer}), and finally explain our method for identifying knowledge neurons (§\ref{sec:Identifying Knowledge neurons}).

\subsection{Multimodal Large Language Models}
\label{subsec:mllm}
MLLMs are generally composed of an image encoder, a linear layer, and an LLM. The image encoder receives an image, divides it into several small image patches, and extracts image features as output. The output of the image encoder is represented as $\mathcal{F} \in R^{P\times d_h}$, where $P$ is the number of image patches and $d_h$ is the hidden size of the image encoder. The image features $\mathcal{F}$ are aligned with text embeddings through a linear layer, and using the hidden size $h$ of the LLM, they are transformed into $\mathcal{F^{\prime}} \in R^{P\times h}$. Finally, the image features are concatenated with text prompt embeddings and fed into the LLM.

\subsubsection{Architecture of MLLMs}
\label{mllm arch}
According to Jin et al. \cite{jin2024efficientmultimodallargelanguage}, a typical MLLM consists of three main modules: a visual encoder, a pretrained language model, and a vision-language projector. The vision-language projector serves as a bridge to align the two modalities, and there are two primary approaches to this alignment.\\
\textbf{MLP-based Approach}\hspace{5mm}
This method integrates image features into the language space using a simple learnable linear projector or a multilayer perceptron (MLP). The MLP consists of alternating linear projectors and nonlinear activation functions. MiniGPT-4, which we use in this study, adopts an MLP-based architecture.\\
\textbf{Attention-based Approach}\hspace{5mm}
BLIP2 \cite{li2023blip} introduced a lightweight Transformer module called Q-Former, which uses learnable query vectors to extract visual features from a frozen vision model. Additionally, Flamingo \cite{alayrac2022flamingo} introduced the Perceiver Resampler, which utilizes learnable latent queries ($Q$) in cross-attention, expanding and integrating image features into key ($K$) and value ($V$) representations. This allows the Transformer output at the corresponding positions of learnable latent queries to function as an aggregated representation of visual features, enabling the standardization of variable-length video frame features into fixed-size representations.\\
\textbf{Other Approaches}\hspace{5mm}
Various other methods, such as CNN-based and Mamba-based approaches, have also been explored \cite{jin2024efficientmultimodallargelanguage}.\\
We focus our analysis on models that adopt the most common MLP-based architecture. This selection is motivated by the constraints of available models for activation heatmap generation (§\ref{sec:Limitations}).

\subsubsection{MiniGPT-4}
\label{minigpt4}
MiniGPT-4 is a model designed to adapt pretrained LLMs, such as Vicuna \cite{vicuna2023} and Llama-2 \cite{touvron2023llama2openfoundation}, to image features output by using a pretrained image encoder. The image encoder used in MiniGPT-4 includes models such as BLIP2 and ViT \cite{dosovitskiy2021imageworth16x16words}.

As described in §\ref{mllm arch}, a bridging module is required between the image encoder and the language model. MiniGPT-4 adopts both MLP-based and attention-based approaches, which are available as separate models. In this study, we use the MLP-based model, which employs ViT as the image encoder and Llama-2 (7B) as the language model. Hereafter, we refer to this version simply as MiniGPT-4.

\subsection{Neurons in Transformer}
\label{sec:Neurons in Transformer}
Typically, a decoder-only Transformer consists of stacked Attention layers and FFN layers. Each layer first applies multi-head self-attention in the Attention layer, followed by residual connections and layer normalization. Then, an FFN layer, composed of two linear layers and an activation function, is applied.

Let $h^l$ be the hidden state of layer $l$, $m^l$ the output of the FFN layer, and $a^l$ the output of the Attention layer. Then, $m^l$ can be expressed as follows:

\begin{equation}
m^l = W^l_{out}\,\sigma \lparen W^l_{in}\lparen a^l + h^{l-1} \rparen \rparen
\label{eq_1}
\end{equation}

where $h^0$ is the input embedding vector, $\sigma$ is the activation function, and $W^l_{in}$ and $W^l_{out}$ correspond to the first and second layers of the FFN, respectively.

Attention is a crucial mechanism in the Transformer, extracting contextually relevant information and knowledge. The FFN layer was initially considered a simple nonlinear transformation. However, in 2020, Geva et al. \cite{DBLP:journals/corr/abs-2012-14913} revealed that the FFN layer also plays an important role in storing knowledge and memory. Their study showed that the FFN layer functions as a key-value transformation for knowledge, where each row of the first weight matrix, $W^l_{in}$, acts as a key, and the columns of $W^l_{out}$ serve as values.

This implies that when a specific key is input, it is activated by the activation function $\sigma$.
The corresponding value is retrieved and then mixed into the intermediate representation through residual connections. Moreover, these value vectors often contain human-interpretable concepts and knowledge. \cite{geva-etal-2022-transformer}

Similarly, Meng et al. \cite{meng2022locating} investigated GPT models and measured the indirect influence of hidden states and activations on responses. They found that factual knowledge is primarily stored in the FFN layers of early model layers. Additionally, Dai et al. \cite{dai-etal-2022-knowledge} demonstrated the existence of knowledge neurons in pretrained Transformer FFN layers through neuron-editing experiments.

Based on these insights, numerous studies have sought to elucidate the knowledge and memory functions of Transformer-based language models. \cite{wang-etal-2022-finding-skill, schwettmann2023multimodalneuronspretrainedtextonly, pan2024findingeditingmultimodalneurons} Consequently, this study, which aims to clarify the location of knowledge in Transformers, also focuses on neurons in the FFN layer.

For simplicity, we define the output of the activation function $\sigma$ as $O^l = \sigma \lparen W^l_{in}\lparen a^l + h^{l-1} \rparen \rparen$ and concatenate the values from all layers to obtain $O$. In this study, we refer to this as the activation value. Here, $O \in \mathbb{R}^{L \times T \times d_f}$, where $L$ represents the number of model layers, $T$ is the length of the input token sequence, and $d_f$ is the intermediate dimension of the FFN layer. We denote the $i$-th neuron of the $l$-th layer as $\lparen Ll, Ui\rparen$ and define the set of all neurons as $A$.

\subsection{Identifying Knowledge Neurons}
\label{sec:Identifying Knowledge neurons}
As described in \ref{sec:Neurons in Transformer}, previous studies have shown that activations in FFN layers are closely related to the knowledge function of Transformers. Based on this, we hypothesize that by observing activation value changes when modifying part of the input features, we can derive the relevance of those features.

Following this hypothesis, this study proposes a method that involves editing input images by removing objects associated with specific knowledge. We then compare neuron activations before and after image modification, using GradCAM for filtering, to determine the relevance of the removed objects (knowledge).

As outlined in \ref{sec:Neurons in Transformer}, we focus on the activation function output $O$ of the FFN layer. Since MiniGPT-4 inputs image features as soft prompts to the LLM, given image input $img$ and text input $text$, the input token sequence (excluding special tokens) is represented as $(img_1, img_2, ..., img_P, text_1, ..., text_T)$.

Restricting the activation values to image tokens, we define $O^{\prime} \in \mathbb{R}^{L \times P \times d_f}$. When targeting knowledge $k$ associated with a specific image $o$, the activation values for image tokens are denoted as $O^{\prime \ k}_{o}$.

In the following sections, we describe the method for identifying neurons based on $O^{\prime \ k}_{o}$ as the starting point.
\paragraph{Activation differences filtering using inpainting}

\label{inpaint}

To generate images with specific knowledge-associated objects removed, this study employs inpainting \cite{Telea01012004}. Given an image $o$ containing an object associated with knowledge $k$, we create an inpainted image $i$ where the corresponding object has been removed. The activation values of image tokens in the inpainted image are denoted as $O^{\prime \,k}_{i}$.

\begin{figure}[t]
\centering
\begin{minipage}[b]{0.49\columnwidth}
    % \centering
    \includegraphics[width=0.9\columnwidth]{images/000000361180.jpg}
\end{minipage}
\begin{minipage}[b]{0.49\columnwidth}
    % \centering
    \includegraphics[width=0.9\columnwidth]{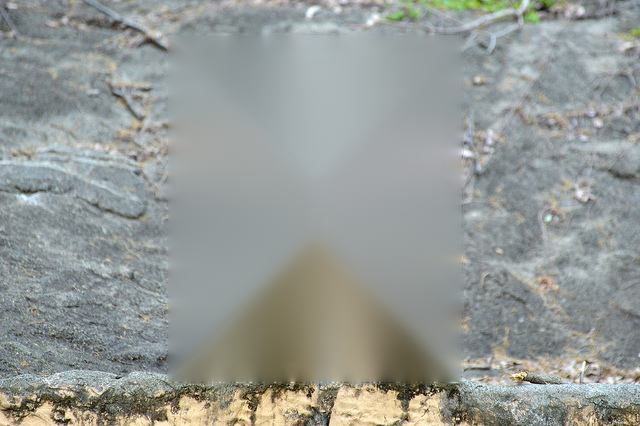}
    
\end{minipage}
\caption{Original and inpainted images}
\label{fig:1}
\end{figure}

Since the inpainting operation is restricted to the region containing the object associated with knowledge $k$, the resulting image appears as if only the object corresponding to knowledge $k$ has been removed. Figure \ref{fig:1} illustrates an example where the knowledge target is "bear," showing both the original and inpainted images.

By computing the difference between activation values, $O^{\prime \,k}_{o} - O^{\prime \,k}_{i}$, we evaluate the neurons most relevant to the target knowledge $k$. Neurons with larger activation differences are considered more strongly associated with knowledge $k$. We define a threshold for activation differences per sample and collect neurons exceeding this threshold into a candidate set $C^k$.

\[
C^k = \lbrace (Ll, Ui)\in A \,|\, V^k[l, p, i] > threshold_{a}^k\rbrace
\]

Here, $p$ denotes an image patch, and $threshold_{a}^k$ represents the activation differences threshold for knowledge $k$. A neuron is selected as a candidate if its activation differences exceed the threshold for at least one patch $P$.

\paragraph{Gradient-based Filtering using GradCAM}
For neurons $(Ll, Ui) \in C^k$ obtained through \ref{inpaint}, we further refine the selection using GradCAM \cite{Selvaraju_2017_ICCV}. This step aims to eliminate neurons activated by factors unrelated to knowledge $k$.

First, we generate an image caption using the text prompt \("The\,\,image\,\,shows\,\,a"\). We confirm that the generated caption contains the correct token $c$ associated with knowledge $k$ and use the preceding token sequence for gradient computation. Given the model’s prediction logit $y_c$ for $c$, the GradCAM value $g_c$ is computed as:

\begin{equation}
g_{c} = O_{o}^{\prime} \frac{\partial y^{c}}{\partial O_{o}^{\prime}}
\end{equation}

Similar to activation value difference filtering, we define a threshold for GradCAM values and extract neurons exceeding the threshold from $C^k$ to construct the final set of neurons associated with knowledge $k$, denoted as $N_k$:

\[
N_k = \lbrace (Ll, Ui)\in C^k \,|\, g_c[l, p, i] > threshold_{g}^k\rbrace
\]

Here, $threshold_{g}^k$ represents the GradCAM threshold for knowledge $k$. 

\section{Experiments}
\label{sec:Experiments}
\subsection{Dataset}
\label{dataset}

In this study, we utilize the 2017 train/val split of the MS COCO dataset \cite{lin2015microsoftcococommonobjects}. Since MS COCO provides bounding box (BBOX) information indicating object regions in images, we leverage this data for the inpainting process described in \ref{inpaint}.

Due to the nature of our method, which relies on activation value differences, the usable data is subject to the following constraints:
the generated caption for the original image must contain the target knowledge;
the generated caption for the inpainted image must not contain the target knowledge;
the inpainting process must remove only the target knowledge.

The MS COCO dataset provides labels for categories such as \{animal, vehicle, food, ...\} and specific objects such as \{horse, bear, airplane, ...\}. Under these constraints, we select three types of knowledge (objects) from each category and sample three instances per object, resulting in a total of 90 data samples, denoted as $D$, for our experiments.

\subsection{Compared Methods}
\label{compared method}

We compare our method with the approach devised by Schwettmann et al. \cite{schwettmann2023multimodalneuronspretrainedtextonly} (abbreviated as Schwettmann's). Their method detects multimodal neurons that map visual features to corresponding text by scoring neurons using gradients, thereby identifying concepts introduced into the model's residual stream.

Additionally, we compare our method with the scoring approach proposed by Pan et al. \cite{pan2024findingeditingmultimodalneurons} (abbreviated as Pan's), which utilizes activation values and FFN layer weights.

Using these methods, neurons are scored and averaged across image token positions. The neurons are then sorted in descending order of score, and the top $|N_k|$ neurons are selected for comparison against our method.

\subsection{Identifying Knowledge Neurons}

\begin{figure}[h]
\centering
\includegraphics[width=0.6\columnwidth]{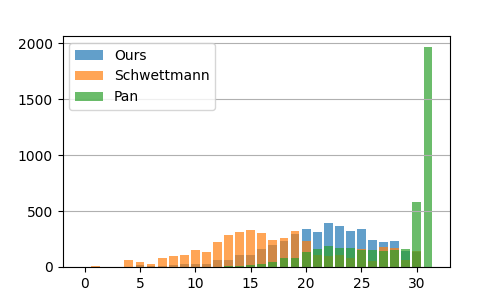}
\caption{Layer distribution of identified neurons}
\label{fig:layer_distribution}
\end{figure}

Following the methodology described in \ref{sec:Identifying Knowledge neurons}, we identify knowledge neurons. Figure \ref{fig:layer_distribution} illustrates the distribution of neurons identified as knowledge neurons across layers for all knowledge instances in the dataset.

Our identified neurons are primarily concentrated in the middle-to-later layers, while Schwettmann's neurons tend to be more concentrated in the earlier layers, and Pan's neurons are heavily concentrated in the final layer.

To conduct a more in-depth analysis, we perform the following experiments.

\subsection{Are the identified knowledge neurons related to the target knowledge?}

We examine whether the identified knowledge neurons are sensitive to the target knowledge in terms of the following:
the influence of knowledge neurons on the target knowledge (\ref{Noise perturbation});
the influence of knowledge neurons on other knowledge (\ref{Noise perturbation});
whether knowledge neurons correspond to visual concepts (\ref{Qualitative evaluation}); and
whether knowledge neurons correspond to textual concepts (\ref{Qualitative evaluation}).

\subsubsection{Image Captioning with Noise Perturbation}
\label{Noise perturbation}

We apply noise to the activation values $O^{\prime}$ of each identified neuron $(Ll, Ui) \in N_k$ across all image token positions and generate captions for all images $I \in D$.

The noise is generated by sampling a vector from a standard normal distribution and scaling it by three different noise levels $n \in \{40, 80, 120\}$, resulting in a noise vector $v \in \mathbb{R}^P$. To account for the randomness of noise sampling, we generate 10 different noise vectors for each noise level and produce 10 captions per level.

The hyperparameters used for caption generation are as follows.

\begin{itemize}
    \item do\_sample = False
    \item temperature = 1.0
    \item num\_beams = 1
    \item max\_new\_tokens = 32
    \item min\_length = 1
    \item top\_p = 0.9
    \item repetition\_penalty = 1
    \item length\_penalty = 1
\end{itemize}

We generate image captions for all datasets used to identify neuron groups, datasets containing the same knowledge objects, and datasets containing different knowledge objects \( \bar{k} \).

\begin{table*}
    \small
    % \hspace{-1.5cm}
    \renewcommand{\arraystretch}{2.5} % 行間の調整
    \captionsetup{justification=centering}
    \begin{tabular}{cccm{10cm}}
        \hline
        $k$ & Image & Target neurons & \multicolumn{1}{c}{\centering Caption} \\
        \hline
        \multirow{4}{*}{\raisebox{-5\height}{bed}} & 
        \multirow{4}{*}{\raisebox{-0.5\height}{\includegraphics[width=3cm]{images/000000249550.jpg}}} & 
        \multicolumn{1}{c}{\centering Original} & The image shows a bedroom with a white bed, a wooden floor, and a floral patterned blanket on the bed. The room has a window with a view of \\
        \cline{3-4}
        & & \multicolumn{1}{c}{\centering Schwettmann's} & The image shows a bedroom with a bed and a bedroom and the city of Bed Bath and Bed Bath and Bed Bath and Bed Bath and Bed Bath and Bed Bath and Bed \\
        \cline{3-4}
        & & \multicolumn{1}{c}{\centering Pan's} & The image shows a bedroom with a white wall, a wooden floor, and a floral patterned quilt on the bed. The room has a window with white curtain \\
        \cline{3-4}
        & & \multicolumn{1}{c}{\centering Ours} & The image shows a room with a wooden floor and walls, and a large window on the wall. hopefully, the room is well-lit and well-ventilated. \\
        \hline
        \multirow{4}{*}{\raisebox{-2.\height}{bear}} & 
        \multirow{4}{*}{\raisebox{-1.\height}{\includegraphics[width=4cm]{images/000000361180.jpg}}} & 
        \multicolumn{1}{c}{\centering Original} & The image shows a brown bear standing on a rocky outcropping, looking out over a body of water. The bear is standing on its hind legs, with its front \\
        \cline{3-4}
        & & \multicolumn{1}{c}{\centering Schwettmann's} & The image shows a RRRRRRRRRRRRRRRRRRRRRRRRRRRRRRR \\
        \cline{3-4}
        & & \multicolumn{1}{c}{\centering Pan's} & The image shows a close-up of a grizzly bear standing on a rocky ledge. hopefully, the image is clear and well-lit. \\
        \cline{3-4}
        & & \multicolumn{1}{c}{\centering Ours} & The image shows a large, dark-haired man standing on a rocky outcropping, looking out over a rocky landscape. everybody is wearing a face mask. \\
        \hline
        \multirow{4}{*}{\raisebox{-5\height}{banana}} & 
        \multirow{4}{*}{\raisebox{-1.0\height}{\includegraphics[width=3cm]{images/000000185802.jpg}}} & 
        \multicolumn{1}{c}{\centering Original} & The image shows a banana sitting on a beach chair in front of a window with a view of the ocean. The banana is wearing sunglasses and has a \\
        \cline{3-4}
        & & \multicolumn{1}{c}{\centering Schwettmann's} & The image shows a banana tree growing in the middle of a room with a window on the left side of the image and a window on the right side of the image. \\
        \cline{3-4}
        & & \multicolumn{1}{c}{\centering Pan's} & The image shows a room with a table and chairs in front of a window that looks out onto the ocean. The table has a small umbrella on it and there are \\
        \cline{3-4}
        & & \multicolumn{1}{c}{\centering Ours} & The image shows a room with a table and chairs in front of a window with a view of the ocean. The table has a vase of flowers on it and a book \\
        \hline
    \end{tabular}
    \caption{
    Example of generated image caption ($n=80$) \\
    Original: Non-noise}
    \label{tab:generated caption}
\end{table*}

% ノイズ生成定量評価
\begin{table*}[t]
    \centering
    % \small
    % \hspace{-2.cm}
    \captionsetup{justification=centering}
    \begin{tabular}{ccccc|ccc|ccc}
        \hline
        \multirow{2}{*}{$k$} & \multirow{2}{*}{$n$} & \multicolumn{3}{c|}{$S_{se}$} & \multicolumn{3}{c|}{$S_{re}$} & \multicolumn{3}{c}{$S_{mean}$}\\
        \cline{3-11}
        &  & Schwettmann's & Pan's  & ours & Schwettmann's & Pan's & ours & Schwettmann's & Pan's & ours \\
        \hline
        
            \multirow{3}{*}{bed} & 40 & 0.4890 & 0.0110 & 0.7447 & 0.7723 & 0.8477 & 0.8320 & 0.6306 & 0.4294 & 0.7883\\
            & 80  & 0.1667 & 0.0000 & 0.9557 & 0.5310 & 0.8227 & 0.7647 & 0.3488 & 0.4113 & 0.8602\\
            & 120  & 0.2333 & 0.0000 & 0.9667 & 0.3370 & 0.7963 & 0.6830 & 0.2852 & 0.3982 & 0.8249\\
            \hline

            \multirow{3}{*}{bear} & 40  & 0.4887 & 0.0000 & 0.1110 & 0.3523 & 0.8083 & 0.7723 & 0.4205 & 0.4042 & 0.4416\\
            & 80  & 0.7553 & 0.1000 & 0.5443 & 0.0087 & 0.7510 & 0.6767 & 0.3820 & 0.4255 & 0.6105\\
            & 120  & 0.9223 & 0.2113 & 0.7887 & 0.0000 & 0.7170 & 0.6077 & 0.4612 & 0.4642 & 0.6982\\
            \hline

        \multirow{3}{*}{banana} & 40  & 0.0000 & 0.2220 & 0.3777 & 0.7637 & 0.8413 & 0.8437 & 0.3819 & 0.5316 & 0.6107\\
    & 80  & 0.3667 & 0.2890 & 0.6113 & 0.4953 & 0.8337 & 0.8143 & 0.4310 & 0.5614 & 0.7128\\
    & 120  & 0.3890 & 0.3110 & 0.7777 & 0.3947 & 0.8333 & 0.7947 & 0.3919 & 0.5722 & 0.7862\\
    \hline
        \hhline{===========}
        Ave. & - & 0.7459 & 0.3723 & \textbf{0.7714} & 0.3742 & \textbf{0.7350} & 0.6957 & 0.5600 & 0.5537 & \textbf{0.7335}\\
        \hline
    \end{tabular}
    \caption{
        Evaluation of image caption generation by noise perturbation.\\
        The bottom line shows the values averaged over all $k$ values in $D$ \\and averaged over the noise levels and averaged across noise levels.
    }
    \label{tab:1}
\end{table*}

\paragraph{Impact on Target and Other Knowledge}
In this study, we evaluate whether \( N_k \) only affects \( k \) by applying noise to \( N_k \) and checking whether \( k \) is absent in the generated caption when an image containing \( k \) is input, and whether \( \bar{k} \) is present in the generated caption when an image containing \( \bar{k} \) is input. If the target knowledge appears in the caption, the score is set to 1; otherwise, it is set to 0.

Since the dataset includes three different images per knowledge instance, three different neuron groups, \( N_k^1, N_k^2, N_k^3 \), are obtained for a given knowledge \( k \). The scores for these neuron groups are averaged and further averaged across noise levels. The score for images containing \( k \) is denoted as \( S_k \in [0,1] \), while the score for images not containing \( k \) is denoted as \( S_{\bar{k}} \in [0,1] \). Since lower values of \( S_k \) indicate that target knowledge was more effectively suppressed, we use \( 1 - S_k \) to measure suppression effectiveness($S_{se}$). Meanwhile, since higher values of \( S_{\bar{k}} \) indicate better retention of other knowledge, we use it directly as a measure of knowledge retention($S_{re}$). The results are shown in Table \ref{tab:1}.

As shown in the results, our method achieves the highest scores in both suppressing target knowledge and retaining other knowledge. While Pan's method achieves the highest retention score, indicating the least impact on other knowledge, its suppression effectiveness is lower, suggesting that Pan's neurons are not strongly related to any specific knowledge. Conversely, Schwettmann's method effectively suppresses target knowledge but has lower retention compared to our method, indicating that it affects other knowledge as well.

These findings suggest that observing activation differences is crucial for identifying neurons strongly associated with specific knowledge.

\paragraph{Comparison with Masked Captions}
Here, we quantitatively compare  the caption of the image masked by inpainting with those generated under noise perturbation, as shown in Table \ref{tab:generated caption}.

The purpose of this evaluation is twofold: to verify whether the grammatical structure of captions is preserved under noise perturbation, and to assess how much information about non-target knowledge present in the image is retained. For example, in Table \ref{tab:generated caption}, Schwettmann's method generates ungrammatical captions for "bear," leading to a lower evaluation score, while our method retains non-target knowledge such as "ocean," "chair," and "window" for "banana," resulting in a higher score.

To achieve these evaluation goals, we adopt three widely used NLP metrics: BLEU \cite{papineni-etal-2002-bleu}, ROUGE \cite{lin-2004-rouge}, and BERTScore \cite{zhang2020bertscoreevaluatingtextgeneration}. The results are summarized in Table \ref{tab:compare to original}.

our method outperforms Schwettmann's method in all evaluation metrics. Higher BLEU and ROUGE scores suggest that our method is capable of generating captions containing more words from the masked captions. Additionally, a higher BERTScore indicates that the captions generated under noise perturbation retain meanings closer to the masked captions.

While Pan's scores surpass ours, Table 2 indicates a lower suppression effect, suggesting minimal differences compared to when non-noise is applied, resulting in smaller deviations from the masked captions. The crucial comparison lies with Schwettmann's, where a notable suppression effect is observed.

Combining these findings with the qualitative examples (Table \ref{tab:generated caption}) and the quantitative evaluation (Table \ref{tab:1}), we conclude that our method surpasses baseline methods in suppressing target knowledge while minimizing deviations from the masked captions and maintaining minimal impact on other knowledge.

\begin{table}[h]
    \centering
    \begin{tabular}{l|c|c|c|c}
        \hline
          & BLEU & ROUGE-1 & ROUGE-L & BERTScore\\
         \hline
         Schwettmann's & 0.3024 & 0.3910 & 0.3529 & 0.8727 \\
         \hline
         Pan's & 0.4615 & 0.5034 & 0.4392 & 0.8971 \\
         \hline
         Ours & 0.4607 & 0.5032 & 0.4374 & 0.8971 \\
         \hline
    \end{tabular}
    \caption{Quantitative comparison with masked captions}
    \label{tab:compare to original}
\end{table}

\subsubsection{Tracing Focus of Neurons in Images and Decoding neurons}
\label{Qualitative evaluation}

In this section, we analyze where the identified neurons respond most strongly in an image and which tokens they represent, providing a qualitative evaluation.

\paragraph{Activation Heatmap}
Following previous studies \cite{schwettmann2023multimodalneuronspretrainedtextonly, pan2024findingeditingmultimodalneurons}, we generate heatmaps to visualize the influence of identified neurons on an image. Schwettmann et al. \cite{schwettmann2023multimodalneuronspretrainedtextonly} created heatmaps using values obtained by applying gradients to activation values, while Pan et al. \cite{pan2024findingeditingmultimodalneurons} directly used activation values. Since our approach primarily relies on activation values for neuron identification, we adopt the same methodology for heatmap generation.

The heatmap generation process is as follows. Let the image size be $d_i \times d_i$. In the case of MiniGPT-4, the number of image tokens input to the LLM after processing by the image encoder is $P=64$. Typically, activation values of image tokens are reshaped into a $\sqrt{P} \times \sqrt{P}$ matrix and then scaled to the image size $d_i \times d_i$ using bilinear interpolation. However, since MiniGPT-4 reduces the number of image tokens input to the LLM by concatenating feature representations of adjacent four patches in the image encoder’s output, we reshape the activation values into a $16 \times 4$ matrix instead of $8 \times 8$ before scaling.

We generate heatmaps by computing the activation values of the top 5 neurons with the highest activation differences among the identified $N_k$ neurons and averaging their activation values, which are then plotted onto the image.

\paragraph{Decoding Neurons}
Following previous studies, we treat each row of the second-layer weight matrix $W_{out}$ of the FFN layer as a vector corresponding to each neuron. By multiplying these vectors with the unembedding matrix, we convert them into a probability distribution over the vocabulary. The top 5 tokens with the highest values are considered as the tokens represented by the corresponding neuron.

\begin{figure}[h]
    \centering
    \includegraphics[width=1\linewidth]{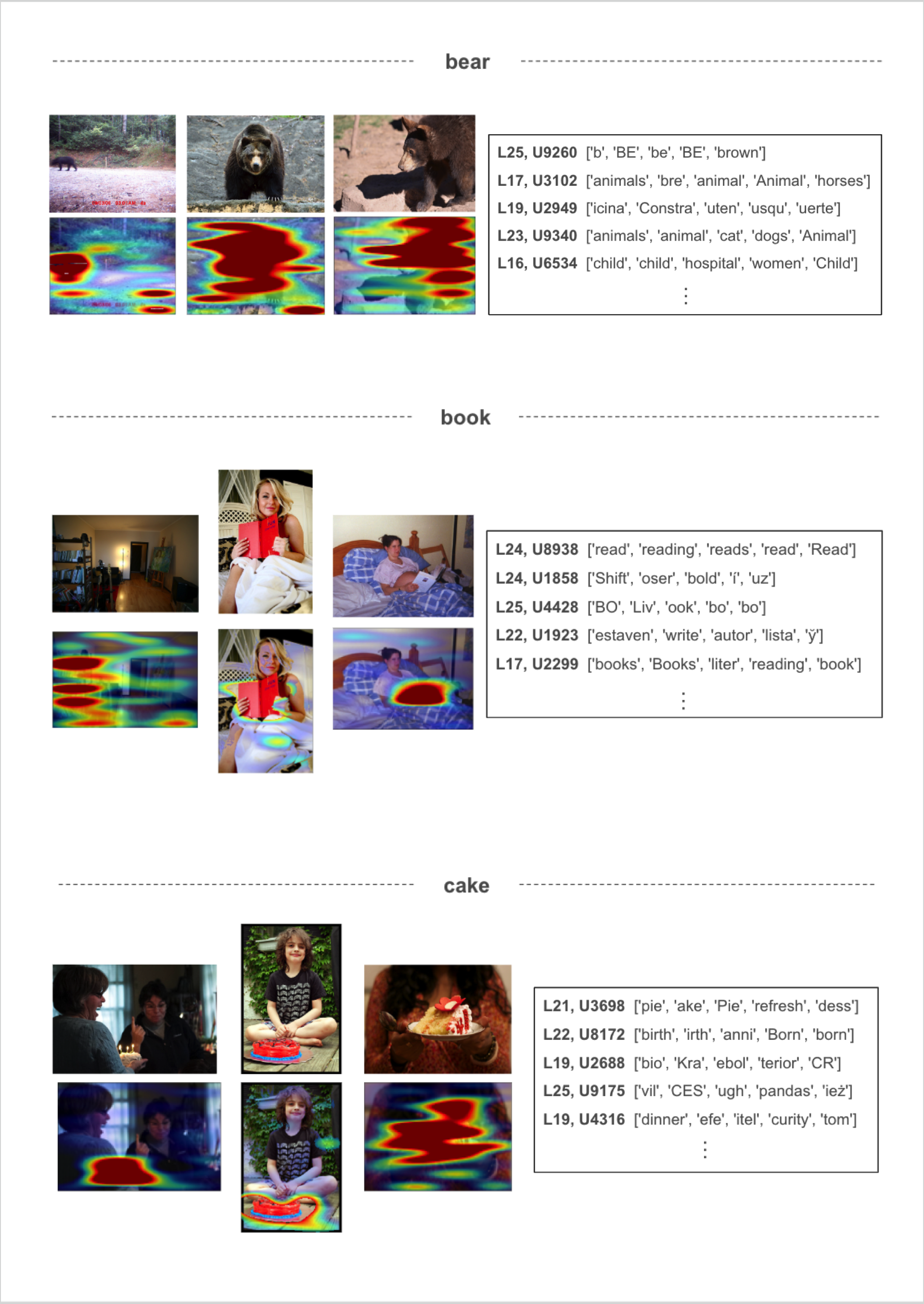}
    \caption{Activation Heatmap and decoding neurons}
    \label{fig:cam}
\end{figure}

Figure \ref{fig:cam} illustrates the results of activation heatmap generation and neuron decoding. The images used to identify neurons for each knowledge category are shown on the far left.

\section{Related work}
\label{sec:Related work}
\subsection{Multimodal Neurons in Pretrained Transformer Language Models}
Schwettmann et al. \cite{schwettmann2023multimodalneuronspretrainedtextonly} explored the existence and role of multimodal neurons in Transformer models pretrained exclusively on text.
Specifically, they combined a frozen Transformer with a self-supervised image encoder and a linear projection layer to analyze how visual information is transformed into textual representations through an image-to-text conversion task.
Their study revealed that visual information is not directly converted into textual representations in the early layers of the model. Instead, they identified "multimodal neurons" in the deeper layers that transform visual representations into corresponding textual expressions.
Furthermore, their experiments demonstrated that these neurons systematically respond to specific visual concepts and have a causal effect on image caption generation.
This research provides insights into how a text-only pretrained model internally processes visual information and utilizes it for text generation, contributing to a deeper understanding of the integration mechanisms between vision and language.

\subsection{Identification and Editing of Multimodal Neurons in Pretrained Transformers}
Pan et al. \cite{pan2024findingeditingmultimodalneurons} focused on the internal mechanisms of MLLMs, particularly the identification and modification of neurons that bridge visual and linguistic concepts.
The authors scored individual neurons by analyzing the final token of textual inputs and proposed a method based on data sensitivity, specificity, and causality to identify multimodal neurons that influence specific visual knowledge.
They introduced a novel approach to efficiently identify key neurons responsible for linking visual and linguistic concepts in caption generation, without requiring gradient computations, which was a limitation of prior methods.
Furthermore, they designed a multimodal knowledge editing technique based on the identified neurons to mitigate sensitive words and hallucinations.
Through extensive quantitative and qualitative experiments grounded in theoretical assumptions, they validated the effectiveness of their proposed method.
This study enhances our understanding of how MLLMs interpret different modalities and integrate cross-modal representations, providing valuable insights for ongoing improvements in both academia and industry.

\section{Limitations}
\label{sec:Limitations}
There are constraints related to the dataset, the model, and the threshold settings.

First, the dataset constraints are threefold: the caption generated when the original image is input must contain the target knowledge; the caption generated when the inpainted image is input must not contain the target knowledge; and the knowledge removed by inpainting must be limited to the target knowledge. These constraints arise from the characteristics of our method, which utilizes activation value differences between the original and inpainted images. If the first constraint is not satisfied due to insufficient model performance, neurons associated with the target knowledge are not initially activated, rendering the activation differences ineffective. Similarly, if the second constraint is not satisfied, neurons related to the target knowledge remain activated even when the inpainted image is input, making the method ineffective. The third constraint is imposed because if the inpainted knowledge object includes non-target knowledge, it becomes impossible to determine which knowledge the activation differences correspond to.

In terms of the model constraints, the only model used in this study is MiniGPT-4 (ViT, LLaMA-2-7B), meaning that the experimental results in \ref{sec:Experiments} may differ for other MLLMs. In addition, the creation of activation heatmaps is only feasible for MLLMs that adopt an MLP-based vision-language projector and input image tokens independently of text input into the LLM. This is because the operation of scaling activation values at each image token position to the original image size depends on a clear correspondence between each input image token and the patches of the input image. Consequently, architectures that use a Q-former or fuse image and text token representations before inputting them into the LLM do not satisfy this dependency, making heatmap generation infeasible.

The threshold constraints require setting two specific thresholds for each knowledge instance: activation value difference and GradCAM. The values used in this study, extracted from a single sample per knowledge instance, are presented in Table \ref{tab:apdx_a_9}. In this study, these values were determined manually; however, future improvements should explore methods for automatic threshold determination.

\section{Conclusion}
\label{sec:Conclusion}
This study takes an initial step toward improving the explainability of Transformer-based text generation models by conducting neuron-level observations using MiniGPT-4, a multimodal large language model (MLLM). We propose a method based on the activation values and gradients of individual neurons at the image token positions in the input sequence.

In text generation experiments with noise perturbation using the MS COCO dataset, our method outperformed Schwettmann’s by approximately 0.17 points and Pan's by approximately 0.18 points in terms of the average score of knowledge suppression effectiveness and retention rate of other knowledge. Furthermore, compared to the masked captions, our method partially achieved higher scores than compared methods across BLEU, ROUGE, and BERTScore metrics. These results quantitatively demonstrate that the neurons identified by our method are more strongly related to the target knowledge than to other knowledge and have minimal impact on other knowledge and grammatical ability.

Additionally, by generating activation heatmaps and decoding neurons, we qualitatively show that the identified neurons are indeed related to the target knowledge.

This study has certain limitations regarding dataset usage and model constraints, highlighting the need for further research and expansion.

%Bibliography
\bibliographystyle{unsrt}  
% \bibliography{references}

\clearpage
\onecolumn

\appendix

\begin{table*}[h]
        \setlength{\tabcolsep}{3pt}
	\small
	\hspace{-0.5cm}
	\renewcommand{\arraystretch}{2.5} % 行間の調整
	\captionsetup{justification=centering}
	\begin{tabular}{cccc|cccc|cccc}
        \hline
        $k$ & image & variable & \multicolumn{1}{c|}{\centering value} & $k$ & image & variable & \multicolumn{1}{c|}{\centering value} & $k$ & image & variable & \multicolumn{1}{c}{\centering value}\\
        \hline
        \multirow{2}{*}{banana} & 
        \multirow{2}{*}{\includegraphics[height=1.5cm]{images/000000185802.jpg}} & 
        $threshold_a$ & 2.5 &
        \multirow{2}{*}{bed} & 
        \multirow{2}{*}{{\includegraphics[height=1.5cm]{images/000000249550.jpg}}} & 
        $threshold_a$ & 1.5 &
        \multirow{2}{*}{bear} & 
        \multirow{2}{*}{{\includegraphics[width=1.5cm]{images/000000361180.jpg}}} & 
        $threshold_a$ & 1.0 \\
        & & $threshold_g$ & 0.0003 & & & $threshold_g$ & 0.0005 & & & $threshold_g$ & 0.0005 \\
        \hline

        \multirow{2}{*}{giraffe} & 
        \multirow{2}{*}{\includegraphics[width=1.5cm]{images/000000245513.jpg}} & 
        \multicolumn{1}{c}{\centering $threshold_a$} & 1.5 &
        \multirow{2}{*}{pizza} & 
        \multirow{2}{*}{\includegraphics[height=1.5cm]{images/000000306700.jpg}} & 
        \multicolumn{1}{c}{\centering $threshold_a$} & 1.5 &
        \multirow{2}{*}{bus} & 
        \multirow{2}{*}{\includegraphics[width=1.5cm]{images/000000249219.jpg}} & 
        \multicolumn{1}{c}{\centering $threshold_a$} & 1.5 \\
        & & $threshold_g$ & 0.0002 & & & $threshold_g$ & 0.0003 & & & $threshold_g$ & 0.0001 \\
        \hline

        \multirow{2}{*}{airplane} & 
        \multirow{2}{*}{\includegraphics[width=1.5cm]{images/000000214205.jpg}} & 
        \multicolumn{1}{c}{\centering $threshold_a$} & 2 &
        \multirow{2}{*}{horse} & 
        \multirow{2}{*}{\includegraphics[width=1.5cm]{images/000000226883.jpg}} & 
        \multicolumn{1}{c}{\centering $threshold_a$} & 1.0 &
        \multirow{2}{*}{refrigerator} & 
        \multirow{2}{*}{\includegraphics[height=1.5cm]{images/000000057238.jpg}} & 
        \multicolumn{1}{c}{\centering $threshold_a$} & 1.25 \\
        & & $threshold_g$ & 0.0 & & & $threshold_g$ & 0.0007 & & & $threshold_g$ & 0.0005 \\
        \hline

        \multirow{2}{*}{toilet} & 
        \multirow{2}{*}{\includegraphics[height=1.5cm]{images/000000042276.jpg}} & 
        \multicolumn{1}{c}{\centering $threshold_a$} & 1.5 &
        \multirow{2}{*}{tennis racket} & 
        \multirow{2}{*}{\includegraphics[height=1.5cm]{images/000000199310.jpg}} & 
        \multicolumn{1}{c}{\centering $threshold_a$} & 1.5 &
        \multirow{2}{*}{fire hydrant} & 
        \multirow{2}{*}{\includegraphics[width=1.5cm]{images/000000384616.jpg}} & 
        \multicolumn{1}{c}{\centering $threshold_a$} & 1.5 \\
        & & $threshold_g$ & 0.0001 & & & $threshold_g$ & 0.0001 & & & $threshold_g$ & 0.0005 \\
        \hline

        \multirow{2}{*}{stop sign} & 
        \multirow{2}{*}{\includegraphics[width=1.5cm]{images/000000079408.jpg}} & 
        \multicolumn{1}{c}{\centering $threshold_a$} & 1.0 &
        \multirow{2}{*}{microwave} & 
        \multirow{2}{*}{\includegraphics[width=1.5cm]{images/000000207538.jpg}} & 
        \multicolumn{1}{c}{\centering $threshold_a$} & 1.0 &
        \multirow{2}{*}{bicycle} & 
        \multirow{2}{*}{\includegraphics[width=1.5cm]{images/000000203317.jpg}} & 
        \multicolumn{1}{c}{\centering $threshold_a$} & 1.5 \\
        & & $threshold_g$ & 0.00075 & & & $threshold_g$ & 0.0002 & & & $threshold_g$ & 0.0004 \\
        \hline

        \multirow{2}{*}{cake} & 
        \multirow{2}{*}{\includegraphics[height=1.5cm]{images/000000102820.jpg}} & 
        \multicolumn{1}{c}{\centering $threshold_a$} & 1.25 &
        \multirow{2}{*}{toaster} & 
        \multirow{2}{*}{\includegraphics[width=1.5cm]{images/000000232348.jpg}} & 
        \multicolumn{1}{c}{\centering $threshold_a$} & 1.0 &
        \multirow{2}{*}{surfboard} & 
        \multirow{2}{*}{\includegraphics[width=1.5cm]{images/000000187362.jpg}} & 
        \multicolumn{1}{c}{\centering $threshold_a$} & 1.5 \\
        & & $threshold_g$ & 0.0003 & & & $threshold_g$ & 0.001 & & & $threshold_g$ & 0.0003 \\
        \hline

        \multirow{2}{*}{snowboard} & 
        \multirow{2}{*}{\includegraphics[width=1.5cm]{images/000000304817.jpg}} & 
        \multicolumn{1}{c}{\centering $threshold_a$} & 1.5 &
        \multirow{2}{*}{traffic light} & 
        \multirow{2}{*}{\includegraphics[height=1.5cm]{images/000000015272.jpg}} & 
        \multicolumn{1}{c}{\centering $threshold_a$} & 2.0 &
        \multirow{2}{*}{keyboard} & 
        \multirow{2}{*}{\includegraphics[width=1.5cm]{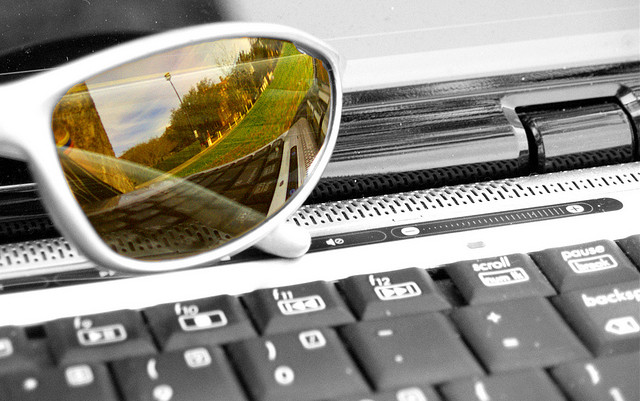}} & 
        \multicolumn{1}{c}{\centering $threshold_a$} & 1.0 \\
        & & $threshold_g$ & 0.002 & & & $threshold_g$ & 0.0002 & & & $threshold_g$ & 0.0005 \\
        \hline

        \multirow{2}{*}{cell phone} & 
        \multirow{2}{*}{\includegraphics[height=1.5cm]{images/000000288391.jpg}} & 
        \multicolumn{1}{c}{\centering $threshold_a$} & 1.5 &
        \multirow{2}{*}{scissors} & 
        \multirow{2}{*}{\includegraphics[width=1.5cm]{images/000000348012.jpg}} & 
        \multicolumn{1}{c}{\centering $threshold_a$} & 1.0 &
        \multirow{2}{*}{book} & 
        \multirow{2}{*}{\includegraphics[width=1.5cm]{images/000000104666.jpg}} & 
        \multicolumn{1}{c}{\centering $threshold_a$} & 1.5 \\
        & & $threshold_g$ & 0.0005 & & & $threshold_g$ & 0.002 & & & $threshold_g$ & 0.0001 \\
        \hline

        \multirow{2}{*}{clock} & 
        \multirow{2}{*}{\includegraphics[height=1.5cm]{images/000000392228.jpg}} & 
        \multicolumn{1}{c}{\centering $threshold_a$} & 2.0 &
        \multirow{2}{*}{suitcase} & 
        \multirow{2}{*}{\includegraphics[height=1.5cm]{images/000000233370.jpg}} & 
        \multicolumn{1}{c}{\centering $threshold_a$} & 1.2 &
        \multirow{2}{*}{backpack} & 
        \multirow{2}{*}{\includegraphics[width=1.5cm]{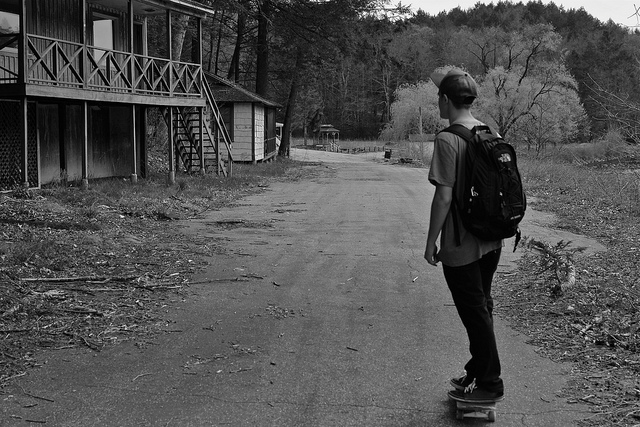}} & 
        \multicolumn{1}{c}{\centering $threshold_a$} & 1.0 \\
        & & $threshold_g$ & 0.0005 & & & $threshold_g$ & 0.0007 & & & $threshold_g$ & 0.001 \\
        \hline
        
        \multirow{2}{*}{umbrella} & 
        \multirow{2}{*}{\includegraphics[height=1.5cm]{images/000000222991.jpg}} & 
        \multicolumn{1}{c}{\centering $threshold_a$} & 0.8 &
        \multirow{2}{*}{laptop} & 
        \multirow{2}{*}{\includegraphics[width=1.5cm]{images/000000243199.jpg}} & 
        \multicolumn{1}{c}{\centering $threshold_a$} & 1 &
        \multirow{2}{*}{vase} & 
        \multirow{2}{*}{\includegraphics[height=1.5cm]{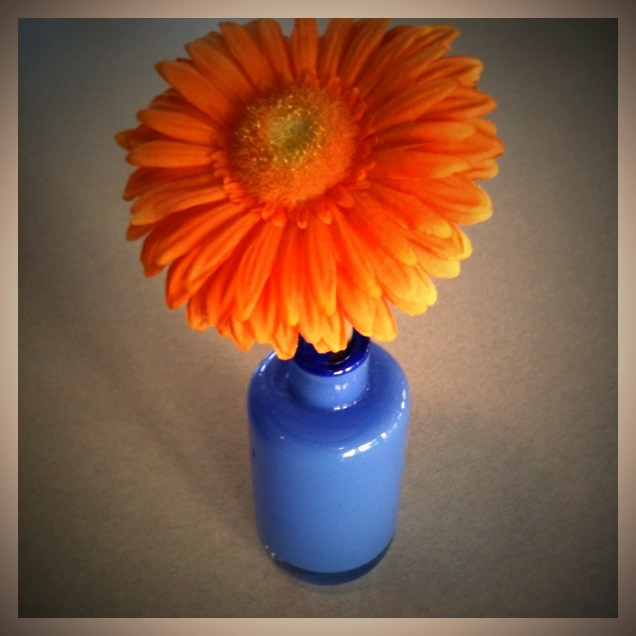}} & 
        \multicolumn{1}{c}{\centering $threshold_a$} & 1 \\
        & & $threshold_g$ & 0.0015 & & & $threshold_g$ & 0.00075 & & & $threshold_g$ & 0.0007 \\
        \hline
        
	\end{tabular}
	\caption{Thresholds of activation differences and GradCAM(one excerpt of each knowledge) 
	}
	\label{tab:apdx_a_9}
\end{table*}

\begin{table*}[h]
    \small
    % \hspace{-1.5cm}
    \renewcommand{\arraystretch}{2.5} % 行間の調整
    \captionsetup{justification=centering}
    \begin{tabular}{cccm{10cm}}
        \hline
        $k$ & Image & Target neurons & \multicolumn{1}{c}{\centering Caption} \\
        \hline
        \multirow{4}{*}{\raisebox{-5\height}{giraffe}} & 
        \multirow{4}{*}{\raisebox{-1.5\height}{\includegraphics[width=3cm]{images/000000245513.jpg}}} & 
        \multicolumn{1}{c}{\centering Original} & The image shows a giraffe standing in a field with trees in the background
The giraffe is standing in a field with trees in the background. The trees are \\
        \cline{3-4}
        & & \multicolumn{1}{c}{\centering Schwettmann's} & The image shows a giraffe standing in a field with trees in the background

The giraffe is standing in a field with trees in the background. The grass is \\
        \cline{3-4}
        & & \multicolumn{1}{c}{\centering Pan's} & The image shows a  \\
        \cline{3-4}
        & & \multicolumn{1}{c}{\centering Ours} & The image shows a mother and a baby in the middle of a field with a fence in the background. The mother is holding the baby in her arms and looking at the camera \\
        \hline
        
        \multirow{4}{*}{\raisebox{-5\height}{pizza}} & 
        \multirow{4}{*}{\raisebox{-1.0\height}{\includegraphics[width=3cm]{images/000000306700.jpg}}} & 
        \multicolumn{1}{c}{\centering Original} & The image shows a woman holding a large pizza in her hands
The woman is wearing a black shirt and has long brown hair. She is smiling and \\
        \cline{3-4}
        & & \multicolumn{1}{c}{\centering Schwettmann's} & The image shows a p p p p p p p p p p p p p p p p p p p p p p p p p p p p p p p p \\
        \cline{3-4}
        & & \multicolumn{1}{c}{\centering Pan's} & The image shows a woman holding a large pizza in her hands  nobody else is in the image
        
        The woman is smiling and appears to be happy with the pizza\\
        \cline{3-4}
        & & \multicolumn{1}{c}{\centering Ours} & The image shows a woman holding a large, cut up, and cut up, and cut up, and cut up, and cut up, and cut up, and cut up, \\
        \hline
    
    \multirow{4}{*}{\raisebox{-5\height}{bus}} & 
        \multirow{4}{*}{\raisebox{-1.5\height}{\includegraphics[width=3cm]{images/000000249219.jpg}}} & 
        \multicolumn{1}{c}{\centering Original} & The image shows a purple and white bus driving down a street with palm trees on either side. The bus has a large number 33 on the side and is travel \\
        \cline{3-4}
        & & \multicolumn{1}{c}{\centering Schwettmann's} & The image shows a street in the 2022 Y Street Project.

The project is a 2-story project that will be used for a variety of purposes \\
        \cline{3-4}
        & & \multicolumn{1}{c}{\centering Pan's} & The image shows a busy street with many cars, trucks, and buses driving by. nobody is in the street. The sky is blue and there are trees \\
        \cline{3-4}
        & & \multicolumn{1}{c}{\centering Ours} & The image shows a street with a few buildings in the background.

The street is lined with trees and has a few cars parked on the side of the road. \\
        \hline
        \multirow{4}{*}{\raisebox{-7.5\height}{airplane}} & 
        \multirow{4}{*}{\raisebox{-1.5\height}{\includegraphics[width=3.5cm]{images/000000214205.jpg}}} & 
        \multicolumn{1}{c}{\centering Original} & The image shows a plane flying in the sky with the moon in the background. \\
        \cline{3-4}
        & & \multicolumn{1}{c}{\centering Schwettmann's} & The image shows a detailed in unison, but the detailed in unison, but the detailed in unison, but the detailed in unison, but the detailed in unison \\
        \cline{3-4}
        & & \multicolumn{1}{c}{\centering Pan's} & The image shows a white background with a black background and a white background with a black background and a white background with a black background and a white background with a black background and a \\
        \cline{3-4}
        & & \multicolumn{1}{c}{\centering Ours} & The image shows a full moon hanging in the sky, with a white building in the foreground. The building is surrounded by trees and has a large window on the side. \\
        \hline
    \end{tabular}
    \caption{
    Example of generated image caption ($n=80$) \\
    Original: Non-noise}
    \label{tab:apdx_generated caption}
\end{table*}

\begin{figure*}[h]
    \centering
    \includegraphics[width=0.75\linewidth]{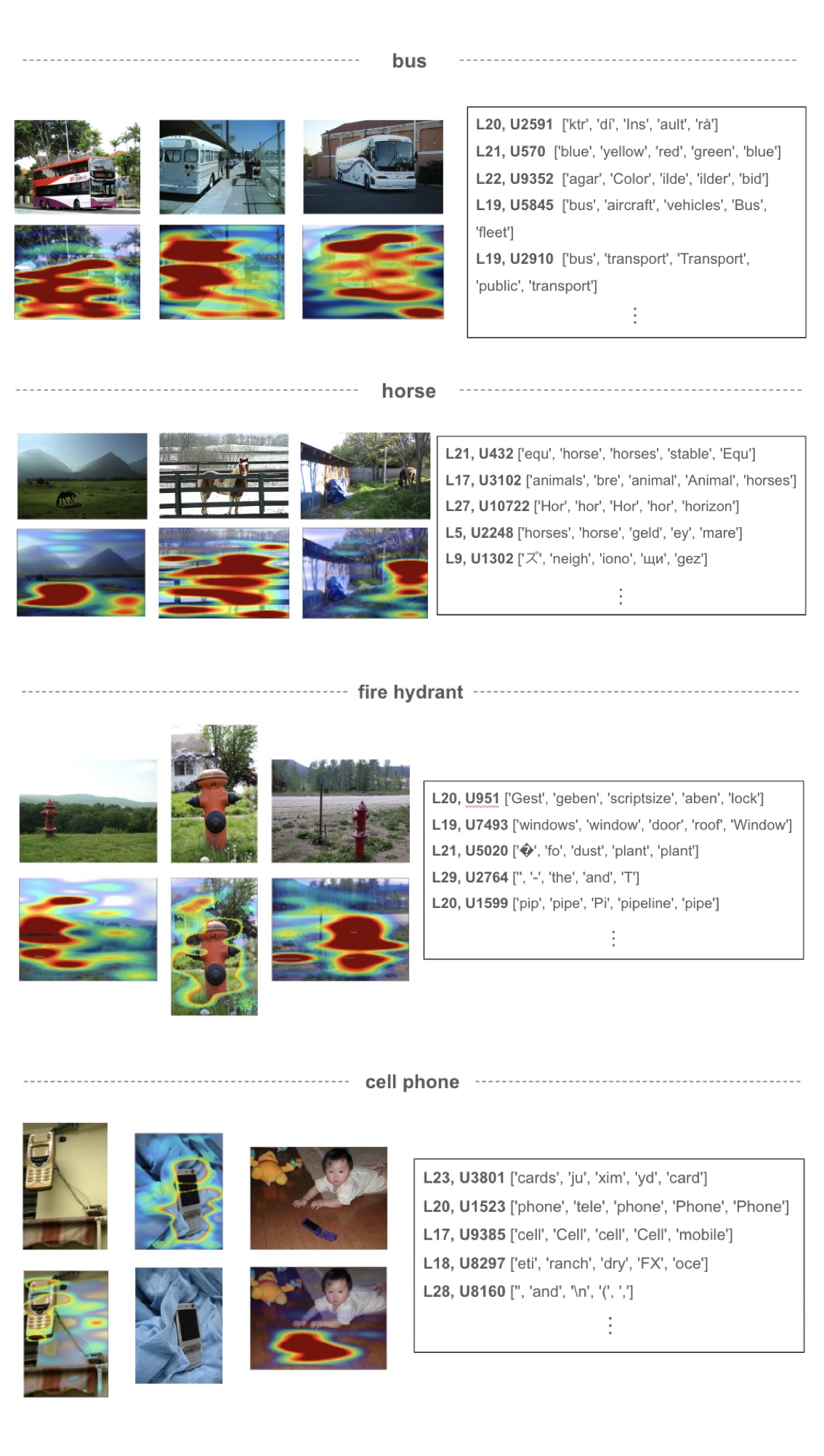}
    \caption{Activation heatmap and Decoding neurons}
    \label{fig:apdx_c_1}
\end{figure*}

\begin{figure*}[h]
    \centering
    \includegraphics[height=0.75\textheight]{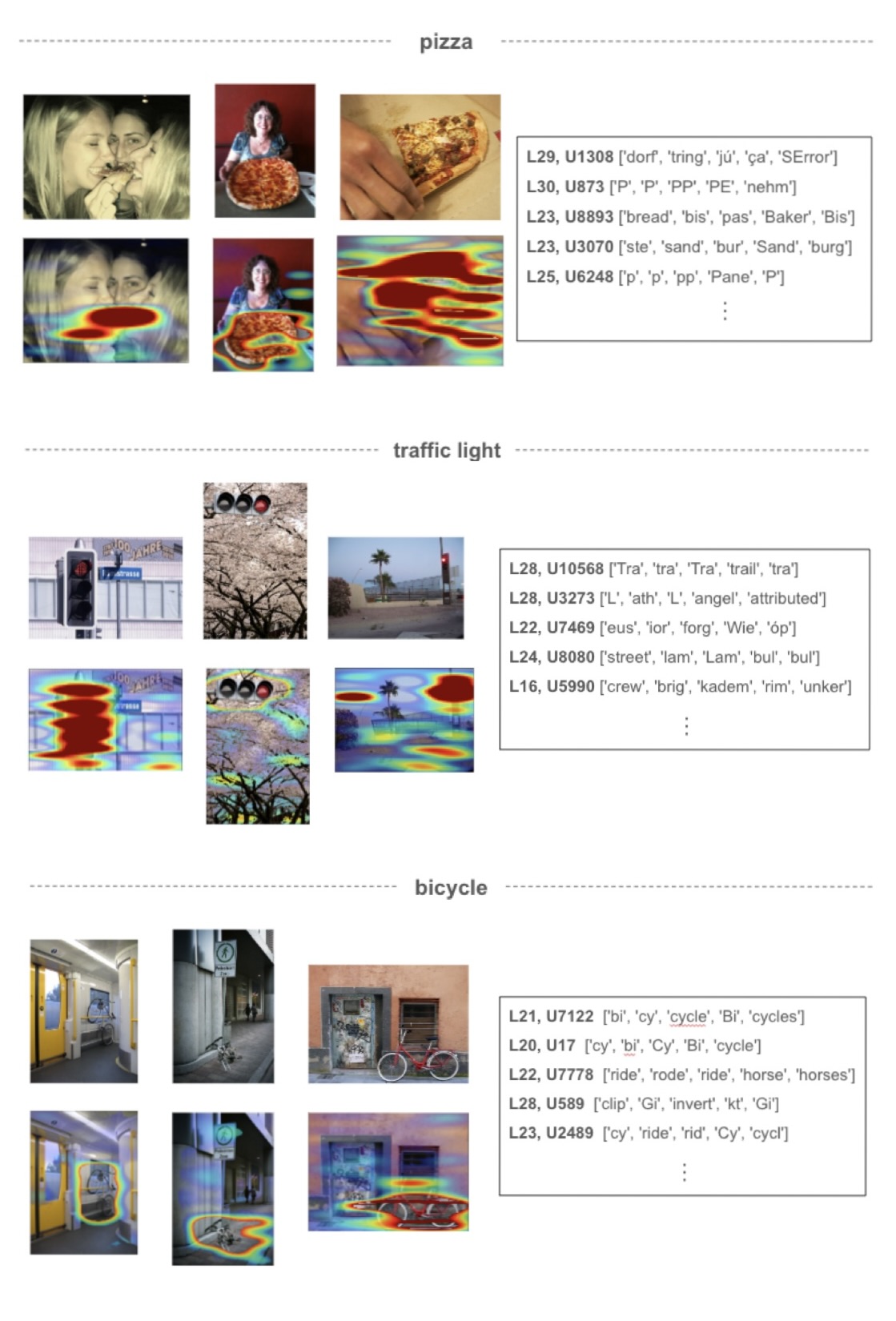}
    \caption{Activation heatmap and Decoding neurons}
    \label{fig:apdx_c_2}
\end{figure*}

\end{document}